\documentclass{ngsm2024} 

\title[Enhancing Transformer RNNs with Multiple Temporal Perspectives]{Enhancing Transformer RNNs with Multiple Temporal Perspectives}

\optauthor{%
 \Name{Razvan-Gabriel Dumitru} \Email{rdumitru@arizona.edu}\\
 \addr University of Arizona, Tucson, AZ, USA
 \AND
 \Name{Darius Peteleaza} \Email{darius.peteleaza@multiversx.com}\\
 \addr MultiversX \newline
 \addr Lucian Blaga University of Sibiu, Sibiu, Romania
 \AND
 \Name{Mihai Surdeanu} \Email{msurdeanu@arizona.edu}\\
 \addr University of Arizona, Tucson, AZ, USA
}

\usepackage{makecell}
\usepackage{microtype}
\usepackage{graphicx}
\usepackage{booktabs} 

\usepackage{hyperref}

\PassOptionsToPackage{table}{xcolor}
\usepackage{colortbl}

\usepackage{amsmath}
\usepackage{amssymb}
\usepackage{mathtools}

\usepackage[capitalize,noabbrev]{cleveref}

\usepackage{times}
\usepackage{latexsym}

\usepackage{amsmath}

\usepackage{caption}
\usepackage{subcaption}

\usepackage{graphicx} 
\usepackage{bm}

\usepackage{array}
\usepackage{tikz}
\usepackage{arydshln}

\newcommand\Tstrut{\rule{0pt}{2.0ex}}         
\newcommand\Bstrut{\rule[-1.0ex]{0pt}{0pt}}   

\usepackage[textsize=tiny]{todonotes}

\begin{document}

\maketitle

\begin{abstract}%
We introduce the concept of multiple temporal perspectives, a novel approach applicable to Recurrent Neural Network (RNN) architectures for enhancing their understanding of sequential data. This method involves maintaining diverse temporal views of previously encountered text, significantly enriching the language models' capacity to interpret context. To show the efficacy of this approach, we incorporate it into the Receptance Weighted Key Value (RWKV) architecture, addressing its inherent challenge of retaining all historical information within a single hidden state. Notably, this improvement is achieved with a minimal increase in the number of parameters --even as little as $0.04\%$ of the original number of parameters. Further, the additional parameters necessary for the multiple temporal perspectives are fine-tuned with minimal computational overhead,
avoiding the need for a full pre-training. The resulting model maintains linear computational complexity during prompt inference, ensuring consistent efficiency across various sequence lengths. The empirical results and ablation studies included in our research validate the effectiveness of our approach, showcasing improved performance across multiple benchmarks. The code, model weights and datasets are open-sourced at \href{https://github.com/RazvanDu/TemporalRNNs}{https://github.com/RazvanDu/TemporalRNNs}.
\end{abstract}




\section{Introduction}

The RWKV (Receptance Weighted Key Value) architecture \cite{peng2023rwkv} bridges the gap between RNNs and Transformers due to its ability to be trained like a Transformer, while offering the inference efficiency of an RNN. The model further stands out for its ability to execute time-parallel processing during training, significantly reducing the computational load. During prompt inference, RWKV ensures linear computational complexity in relation to the sequence length in its sequential decoding mode, akin to conventional RNNs, thus providing consistent efficiency across various sequence lengths. 

Here we address the above limitation with a novel idea in the realm of RNN architectures: {\em multiple temporal perspectives}. At a high level, our method maintains multiple temporal views of the previously seen text, thereby enriching the language model's understanding of sequential data. 
Figure~\ref{fig:example_weights} shows an actual example that highlights how the different perspectives are used during decoding.

The main contributions of our paper are:
\begin{itemize}
    \item A novel approach that maintains multiple temporal perspectives within the RWKV architecture, enhancing its capacity to process and interpret complex language data.
    \item Empirical demonstration of the model's capacity to learn different perspectives from a limited amount of data with a minimal (even as little as $0.04\%$) increase in the number parameters.
    Despite these constrained settings, we show that that our approach outperforms the original RWKV architecture on several benchmarks.
    \item An ablation analysis which indicates the importance of maintaining multiple perspectives as well as the significance of their careful integration. 
\end{itemize}

\section{Related Work}

Addressing the computational demands of Transformer-based architectures has been a fundamental area of research in recent years. Efforts have centered around optimizing the attention mechanism, a cornerstone of Transformer efficiency but also a source of its computational complexity. Innovations in this domain have led to various Transformer adaptations~\cite{beltagy2020longformer,guo-etal-2022-longt5}, which seek to streamline the attention mechanism to reduce computational load. These adaptations include introducing sparse attention patterns \cite{mostafazadeh-etal-2016-corpus,10.1145/3530811} and formulating methods to compute the attention matrix in a more resource-efficient manner \cite{dao2022flashattention,ma2023mega,dao2023flashattention2}.

State space models (SSMs) have been gaining traction in NLP for their ability to model sequential data efficiently. The S4 model \cite{gu2022efficiently} is a notable example, demonstrating the effectiveness of SSMs in handling long-range dependencies within text. Variants of S4, such as those explored by \cite{fu2023hungry,poli2023hyena} and \cite{gu2023mamba}, further refine this approach, offering improvements in scalability and performance. SSMs represent a shift from traditional recurrent architectures, providing a framework that is well-suited for modeling complex temporal dynamics in language data.


\section{Proposed Approach}

\begin{figure*}[ht]
\vskip 0.2in
\begin{center}
\centerline{\includegraphics[width=1\textwidth]{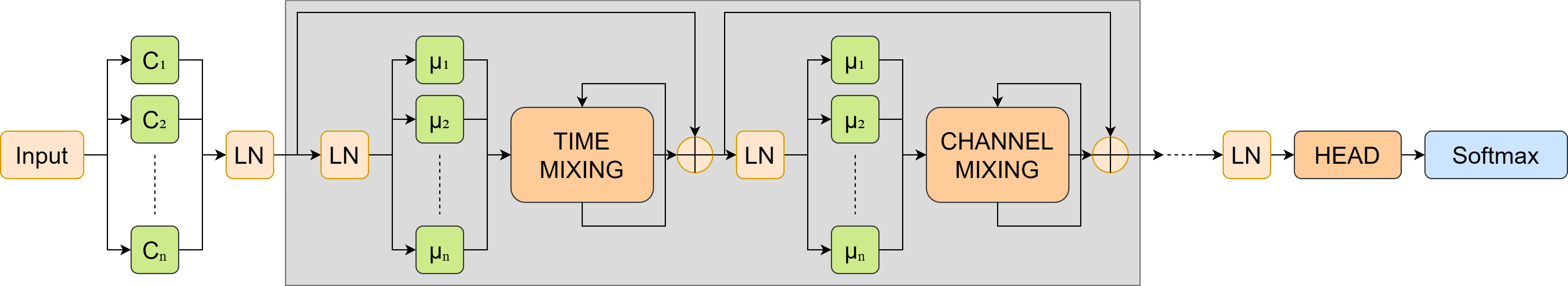}}
\caption{\footnotesize Proposed architecture with multiple temporal perspectives. The green blocks represent the newly added perspective-specific elements. Each input copy (\(C_1\) - \(C_n\)) corresponds to one perspective, processed by its dedicated temporal component (\(\mu_1\) - \(\mu_n\)), used for training. LN denotes  Layer Normalization components. The TIME MIXING, CHANNEL MIXING and HEAD components are described in detail in Figure~\ref{fig:time_mixing}, Figure~\ref{fig:channel_mixing} and in Section \ref{sec:perspectives}. }
\label{fig:model_architecture}
\end{center}
\vskip -0.2in
\end{figure*}

\subsection{Multiple Temporal Perspectives}
\label{sec:perspectives}

Unlike RWKV’s \cite{peng2023rwkv} single state per channel mixing block, our method employs $n$ states in each block, enabling the model to process data from multiple temporal contexts in parallel. This development extends to the time mixing blocks as well. RWKV has one state in each time mixing block, complemented by four additional blocks for computing attention. We expand this by having $n$ such blocks operating in parallel. \textit{This parallel structure allows each block to independently process different segments of temporal data, leading to a more comprehensive understanding of the input sequence, as shown in Figure \ref{fig:example_weights}.} Mathematically, the time-mixing blocks are described by equations \ref{eq:r_t_persp}, \ref{eq:k_t_persp}, and \ref{eq:v_t_persp}. Similarly the channel-mixing blocks are described by Equations \ref{eq:r_t_persp} and \ref{eq:k_t_persp}:

{
\setlength{\abovedisplayskip}{-10pt} 
\setlength{\belowdisplayskip}{0pt} 
\begin{flalign}
&r_t^{(i)} = W_r \cdot (\mu_r^{(i)} \cdot x_t^{(i)} + (1 - \mu_r^{(i)}) \cdot x_{t-1}^{(i)}) \label{eq:r_t_persp} && \\
&k_t^{(i)} = W_k \cdot (\mu_k^{(i)} \cdot x_t^{(i)} + (1 - \mu_k^{(i)}) \cdot x_{t-1}^{(i)}) \label{eq:k_t_persp} && \\
&v_t^{(i)} = W_v \cdot (\mu_v^{(i)} \cdot x_t^{(i)} + (1 - \mu_v^{(i)}) \cdot x_{t-1}^{(i)}) \label{eq:v_t_persp} &&
\end{flalign}
}where $i$ iterates over all available perspectives.

Each perspective $i$ uses replicated temporal components (\(\mu_r^{(i)}\), \(\mu_k^{(i)}\), \(\mu_v^{(i)}\)) that enable the model to combine the current time step (\(x_t^{(i)}\)) with the previous one (\(x_{t-1}^{(i)}\)). The temporal components $\mu$ are essential as they are the ones being customized during our fine-tuning process. The positional weight decay vectors (\(W_r\), \(W_k\), \(W_v\)) are shared across perspectives, which is critical in managing memory usage efficiently due to their substantial memory footprint. The duplication of (\(x_t^{(i)}\)) inputs across each perspective reflects the model's adaptive response to the introduction of multiple temporal viewpoints, directly affecting the inputs processed through each perspective.


\subsection{Perspective Aggregation Strategies}
\label{sec:aggregation}

We explored three distinct methods for combining the outputs from the multiple temporal perspectives incorporated into our model:
average, transformer-like, and a novel method developed by us that is a weighted average with weights learned from data. These aggregation techniques implement the HEAD component in Figure~\ref{fig:model_architecture}. Due to space limitations, they are described in detail in the Appendix (Section \ref{sec:perspective_aggregation_strategies}). 

\section{Experiments}

\subsection{Experimental Setup}

In our experimental framework, we used the English subset of the Wikipedia dataset (dump of 1\textsuperscript{st} of March 2022) \cite{wikidump}, for training our multiple temporal perspectives. We start from pre-trained RWKV-v4 models \cite{RWKVWeights}, which were originally trained on The Pile dataset \cite{gao2020pile}. It should be noted that the Wikipedia dataset is included as part of The Pile, the dataset that RWKV-v4 was pre-trained on, so no additional data was used for training beyond this pre-existing corpus.

\subsection{Model Comparison}

\begin{table*}[h]
\caption{\footnotesize Performance of our method on five datasets compared against three other comparable language models. Green cells (and italic font) indicate results where our model exceeded the performance of the default RWKV-v4 architecture; purple cells denote results that are within the margin of error compared to RWKV-v4; and red cells highlight instances where our model underperformed relative to RWKV-v4. For clarity and emphasis, the best results per dataset among all compared models are highlighted in bold. This color scheme is designed to be accessible to individuals with color vision deficiencies.}
\vskip 0.15in
\resizebox{\textwidth}{!}{
\centering
\begin{tabular}{lllllllll}
\Xhline{4\arrayrulewidth}
\Tstrut
Model & \textbf{Param} & \textbf{LAMBADA} & \textbf{LAMBADA} & \textbf{ARC-E} & \textbf{WinoGrande} & \textbf{HellaSwag} & \textbf{PIQA} \\
      &               B ↓ &     ppl ↓ & acc ↑ & acc ↑ & acc ↑ & acc\_norm ↑ & acc ↑ \Bstrut\\
\hline
\Tstrut
Ours & 0.17 & \cellcolor{green!10}\textit{29.25}$\pm{0.24}$ & \cellcolor{green!10}\textit{34.11}$\pm{0.15}$ & \cellcolor{green!10}\textit{\textbf{47.41}}$\pm{0.27}$ & \cellcolor{green!10}\textit{51.56}$\pm{0.36}$ & \cellcolor{blue!10}\textit{\textbf{32.37}}$\pm{0.07}$ & \cellcolor{red!10}64.56$\pm{0.08}$      \\
RWKV-v4   & 0.17 & 30.55 & 32.91 & 46.88 & 50.82 & 32.36 & \textit{\textbf{64.79}}        \\
\hdashline[2pt/6pt] 
Pythia   & 0.16 & \textbf{24.38} & \textbf{38.97} & 45.12 & \textbf{52.01} & 31.63 & 62.68      \\
GPT-Neo   & 0.16 & 30.27 & 37.36 & 43.73 & 50.43 & 30.42 & 63.06      \Bstrut\\
\hline
\Tstrut
Ours & 0.43 & \cellcolor{green!10}\textit{12.56}$\pm{0.08}$ & \cellcolor{green!10}\textit{46.53}$\pm{0.1}$ & \cellcolor{green!10}\textbf{\textit{53.04}}$\pm{0.28}$ & \cellcolor{blue!10}52.03$\pm{0.12}$ & \cellcolor{blue!10}\textit{\textbf{40.72}}$\pm{0.08}$ & \cellcolor{blue!10}67.57$\pm{0.39}$      \\
RWKV-v4   & 0.43 & 13.04 & 46.05 & 52.65 & \textit{52.09} & \textbf{\textit{40.72}} & \textbf{\textit{67.84}}        \\
\hdashline[2pt/6pt] 
Pythia   & 0.40 & \textbf{11.58} & \textbf{50.44} & 50.38 & \textbf{53.35} & 39.10 & 66.70      \\
GPT-Neo   & 0.40 & 13.88 & 47.29 & 48.91 & 51.14 & 37.64 & 65.07      \Bstrut\\
\hline
\Tstrut
Ours & 1.5 & \cellcolor{green!10}\textit{6.85}$\pm{0.01}$ & \cellcolor{blue!10}57.28$\pm{0.16}$ & \cellcolor{green!10}\textbf{\textit{60.94}}$\pm{0.27}$ & \cellcolor{green!10}\textit{55.38}$\pm{0.16}$ & \cellcolor{blue!10}52.72$\pm{0.14}$ & \cellcolor{red!10}72.01$\pm{0.06}$      \\
RWKV-v4   & 1.5 & 6.91 & \textit{57.36} & 60.52 & 54.53 & \textbf{\textit{52.79}} & \textbf{\textit{72.14}}        \\
\hdashline[2pt/6pt] 
Pythia   & 1.4 & \textbf{6.58} & \textbf{60.43} & 57.74 & \textbf{56.51} & 50.82 & 71.11      \\
GPT-Neo   & 1.4 & 7.5 & 57.25 & 56.19 & 54.93 & 48.94 & 71.16      \Bstrut\\
\Xhline{4\arrayrulewidth}
\end{tabular}
}
\label{table:model_comparison}
\end{table*}

In this subsection of our experiments, we present a detailed analysis comparing the performance of our enhanced RWKV-v4 model with several established models, including the default RWKV-v4, Pythia \cite{biderman2023pythia}, and GPT-Neo \cite{black2022gptneoxb}. Our evaluation uses a suite of popular and challenging benchmarks, namely LAMBADA \cite{paperno2016lambada}, ARC-Easy \cite{Clark2018ThinkYH}, WinoGrande \cite{10.1145/3474381}, HellaSwag \cite{zellers-etal-2019-hellaswag}, and PIQA \cite{Bisk_Zellers_Le_bras_Gao_Choi_2020}.

Table~\ref{table:model_comparison} summarizes our results.
The table shows that our method outperforms the RWKV-v4 model by a statistically significant margin on 10 out of 18 configurations and is comparable on 6 out of the remaining 8 settings. Notably, the improvements in model performance are achieved with a minimal increase in the number of parameters (Table~\ref{table:number_parameters}), underscoring the efficiency of our approach. Further, we ensure that our enhancements improve model capabilities without incurring a substantial computational cost, reflecting a significant advancement in model optimization techniques. For example, using a single NVIDIA RTX 4090 GPU, training one mini-epoch on the smallest model took only approximately 30 minutes.

All in all, this comparative analysis highlights the competitive advantage of our method, which reinforces the potential of our multiple temporal perspectives approach to enhance RNN-based models in a variety of complex NLP tasks.

\subsection{Ablation Studies}

To further explore the the effects of various components and design choices within our approach we also performed ablation studies, which can be found in the Appendix (Section \ref{sec:ablation_studies}). 

\section{Conclusion}

Our research introduced an advancement to RNN architectures, namely multiple temporal perspectives. Our contribution enhances sequential data interpretation with minimal impact on model complexity. 
Our design features parallelizable temporal perspectives, reminiscent of the multi-head attention mechanism in transformers \cite{vaswani2023attention}, and by having these perspectives share most of the parameters, we achieve a minimal parameter increase ($< 0.1\%$), ensuring our architecture is both efficient and streamlined, maintaining a balance between complexity and functionality. The efficient parameter sharing is what truly differentiates our paper from a Mixture of Experts (MoEs) approach where each smaller model has its own complete set of parameters.

We apply our idea to the RWKV architecture \cite{peng2023rwkv}, a recent hybrid approach that trains like a transformer, but performs inference in linear time similar to a RNN.
By applying our multiple temporal perspectives to RWKV, we effectively address the challenge of historical information retention. 
Our empirical results demonstrate enhanced performance across five challenging benchmarks, affirming the potential of this approach to augment RNN capabilities while maintaining linear computational complexity during inference.

\section*{Appendix}
\label{sec:appendix}

\subsection*{Example of Perspective Weights}

\begin{figure}[ht]
\vskip 0.2in
\begin{center}
\centerline{\includegraphics[width=0.5\textwidth]{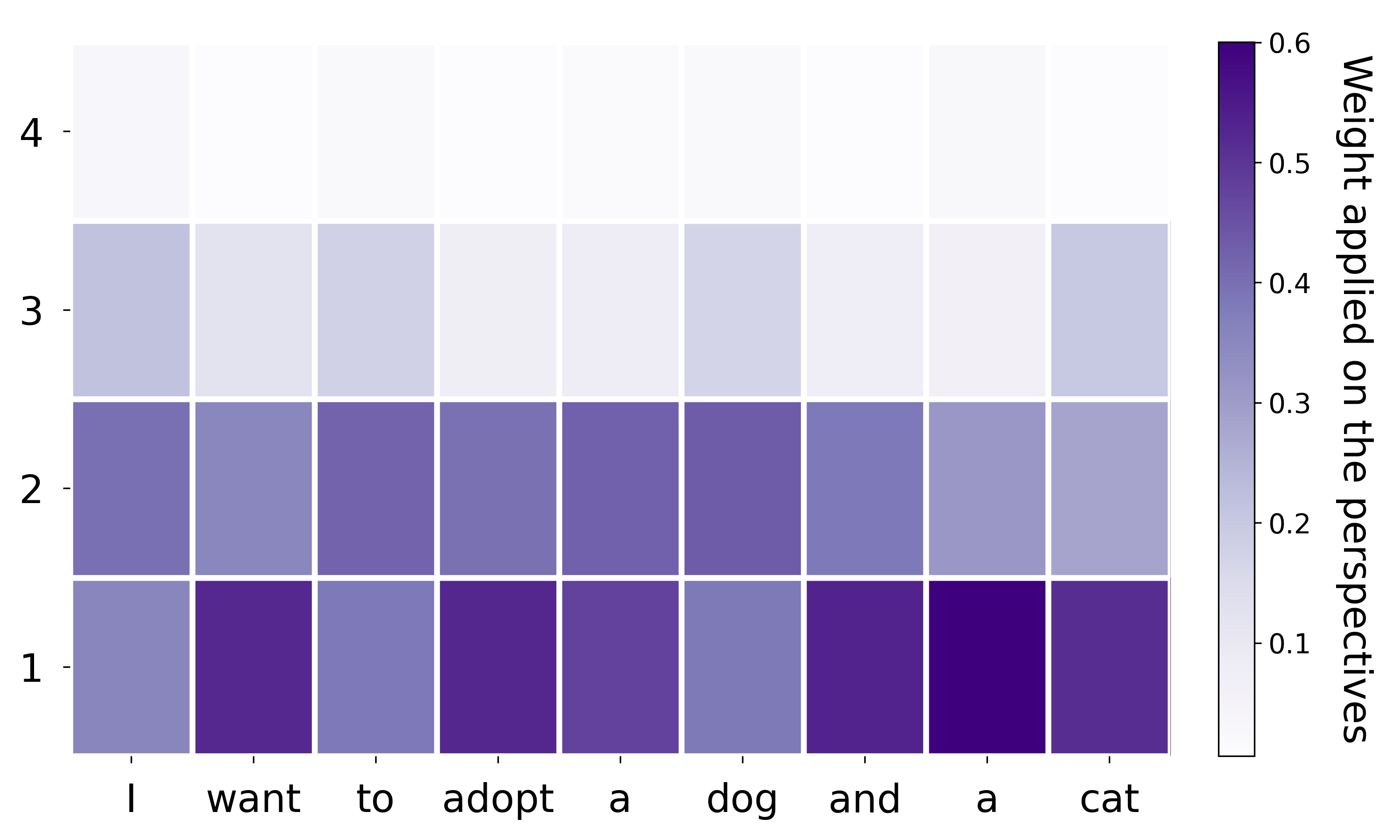}}
\caption{\footnotesize Runtime example of the proposed RNN decoder with four different temporal perspectives. The X axis shows a sentence that was decoded left-to-right. Column $i$ shows the distribution of importance weights assigned to the four temporal perspectives when decoding the token at position $i$. For example, when decoding the word ``dog,'' the model prioritizes Perspective 2, followed by 1, 3, and 4. Perspectives 1 and 2 have higher weights in most cases, reflecting their practicality for capturing commonly occurring syntactic patterns, while Perspectives 3 and 4 are applied in contextually specific instances where unique temporal considerations are required.}
\label{fig:example_weights}
\end{center}
\vskip -0.2in
\end{figure}

\subsection*{Architectural Components}

\begin{figure}[h!]
\vskip 0.2in
\centering
\begin{minipage}[b]{0.48\linewidth}
    \centering
    \includegraphics[width=\linewidth]{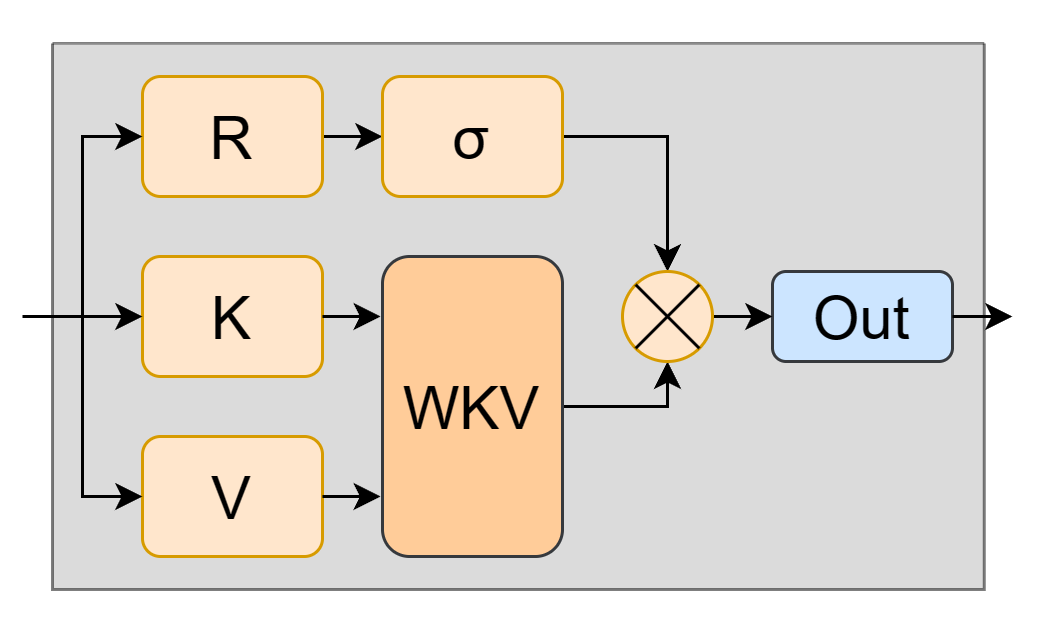}
    \caption{\footnotesize Time mixing}
    \label{fig:time_mixing}
\end{minipage}
\hfill
\begin{minipage}[b]{0.4\linewidth}
    \centering
    \includegraphics[width=\linewidth]{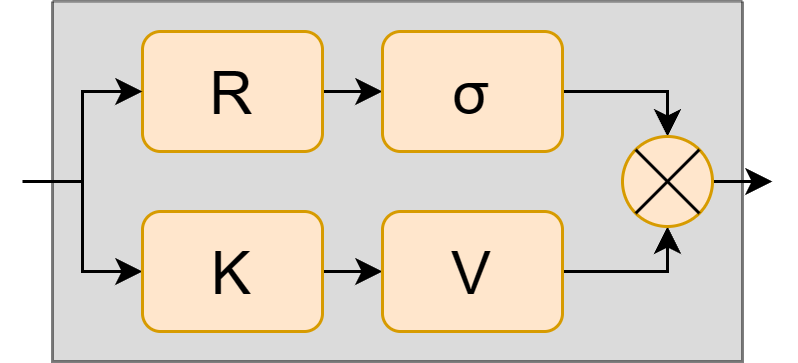}
    \caption{\footnotesize Channel mixing}
    \label{fig:channel_mixing}
\end{minipage}
\vskip -0.2in
\end{figure}

\subsection*{Perspective Aggregation Strategies}
\label{sec:perspective_aggregation_strategies}

\subsubsection*{Average aggregation}



\setlength{\abovedisplayskip}{5pt} 
\setlength{\belowdisplayskip}{5pt} 
\begin{align}
&O_{avg} = head(\frac{1}{n} \sum_{i=1}^{n} p_i) \label{eq:avg_eq} &&
\end{align}

\begin{figure}[h!]
\vskip 0.2in
    \centering
    \centerline{\includegraphics[width=0.36\textwidth]{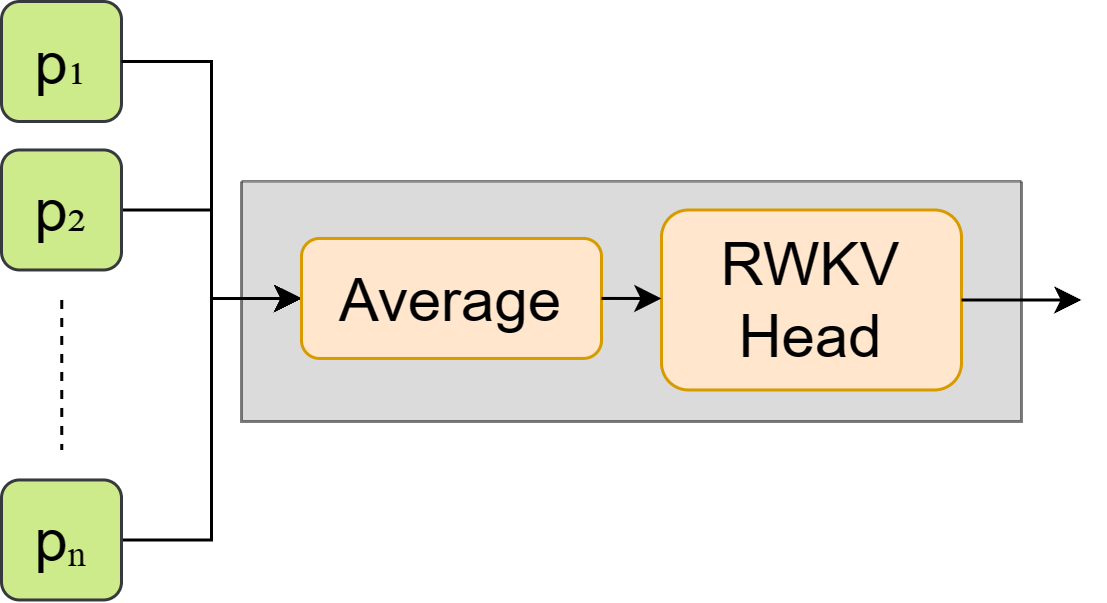}}
    \caption{\footnotesize Simple average aggregation mechanism}
    \label{fig:simple_average}
\end{figure}

The first method we tested was a simple average aggregation approach (Figure~\ref{fig:simple_average}). In this method, the outputs of all temporal perspectives are treated equally, regardless of the specific token being processed. This lack of data dependency means that each perspective contributes uniformly to the final output. While conceptually straightforward, our empirical experiments showed that this approach posed a significant challenge in terms of learning. The neural network had to work harder to discern meaningful patterns and information from equally weighted perspectives, which in some cases could lead to less efficient learning. The equation \ref{eq:avg_eq} describes the mechanism in detail. In all figures, the green blocks (\(p_1\) - \(p_n\)) represent the temporal perspectives.

\subsubsection*{Transformer-like aggregation}

\setlength{\abovedisplayskip}{5pt} 
\setlength{\belowdisplayskip}{5pt} 
\begin{align}
&O_{\text{trans}} = \text{head}\left(W \cdot \text{Concat}(p_1, p_2, \ldots, p_n) + b\right) \text{, where} \nonumber \\
&W \in \mathbb{R}^{\text{embd} \times (n \cdot \text{embd})} \text{ and} \quad b \in \mathbb{R}^{\text{embd}} \label{eq:transf_eq}
\end{align}

\begin{figure}[h!]
    \centering
    \centerline{\includegraphics[width=0.48\textwidth]{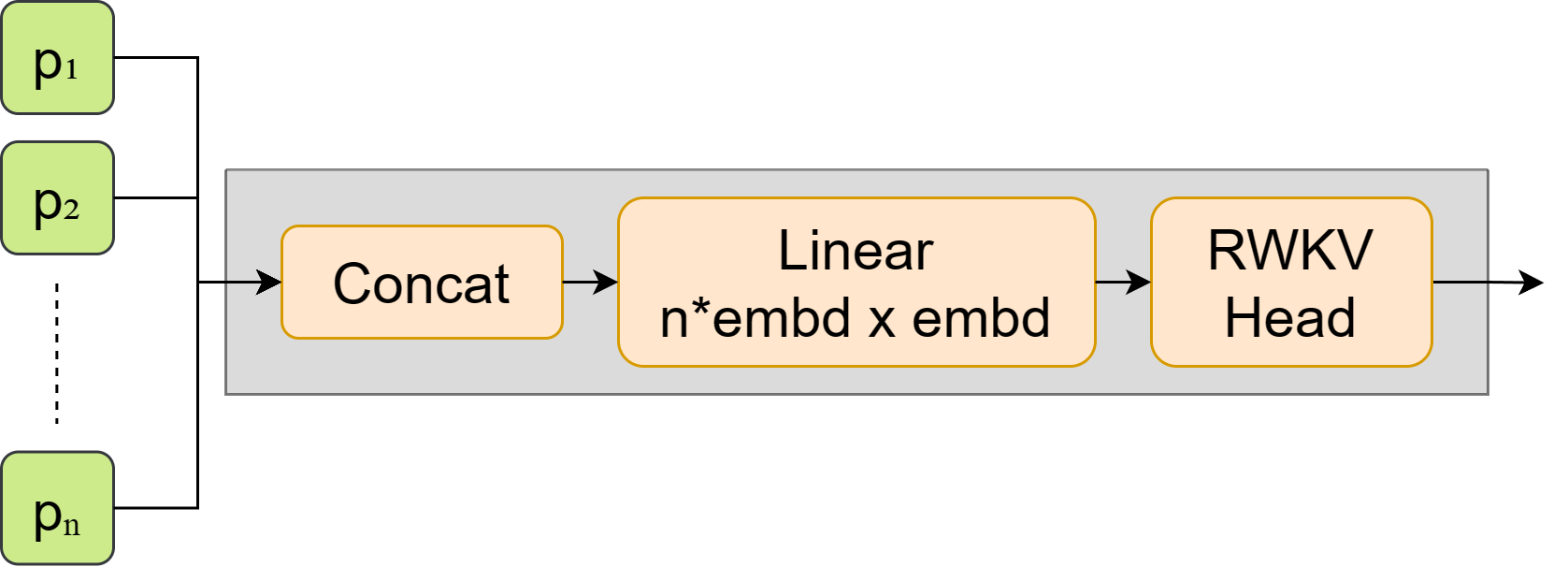}}
    \caption{\footnotesize Transformer-like aggregation mechanism}
    \label{fig:transformer_approach}
\end{figure}

The second method we explored was inspired by the head aggregation strategy in the transformer architecture (Figure~\ref{fig:transformer_approach}). This method introduces data dependency, addressing a key limitation of the average method. It works by taking a concatenation of all embeddings produced by the different perspectives and then employing an MLP-based approach to reduce their size to a single embedding, which is then used in downstream computations. However, this method comes with a significant drawback: it adds a considerable number of extra parameters to the model. Our experiments revealed that this approach was time-consuming in terms of learning. Even with sufficient training time, the model struggled to learn effectively, making this method less feasible for practical applications. The equation \ref{eq:transf_eq} describes the mechanism in detail.

\subsubsection*{Our original aggregation}

\begin{align}
&O_{\text{softmax}} = \text{Softmax}\left(W \cdot (\frac{1}{n} \sum_{i=1}^{n} p_i) + b\right) \text{, where } \nonumber \\
&W \in \mathbb{R}^{n \times \text{embd}} \text{ and } b \in \mathbb{R}^{n} \label{eq:temp_eq} \\
&O_{\text{ours}} = \sum_{i=1}^{n} O_{\text{softmax}, i} \cdot head(p_i) \label{eq:final_eq}
\end{align}

\begin{figure}[h!]
    \centering
    \centerline{\includegraphics[width=0.6\textwidth]{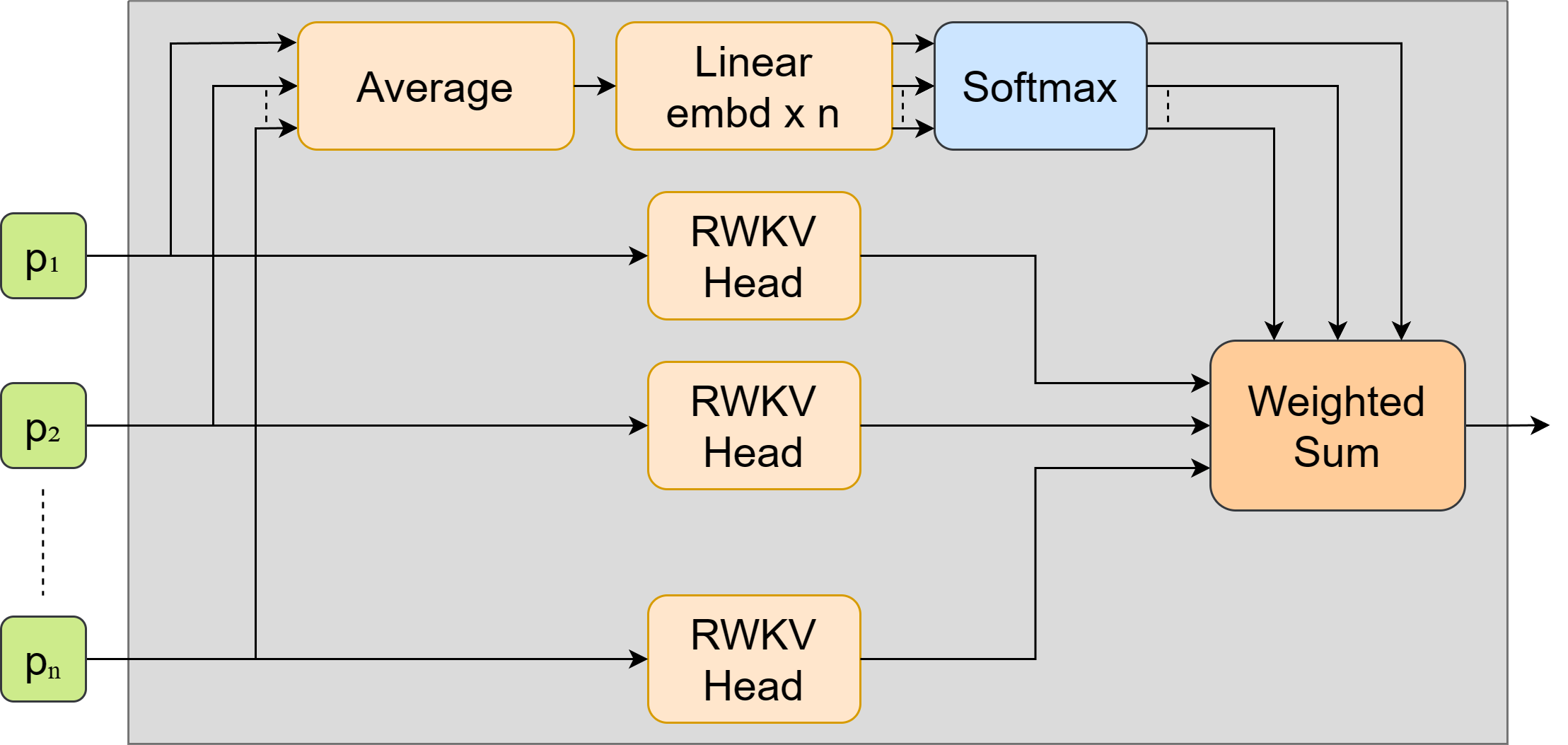}}
    \caption{\footnotesize Our approach, weighted average with learned weights aggregation mechanism}
    \label{fig:our_combination}
\vskip -0.2in
\end{figure}

Our original aggregation method (Figure~\ref{fig:our_combination}) also incorporates data dependency, thus avoiding the shortcomings of the simple average approach introduced above. Our approach begins with an averaging component; this is followed by a context-dependent weighting that contains a linear layer, which evaluates and assigns relevance to each perspective for a given context. The final output is then constructed by combining the outputs from all perspectives, with the contribution of each perspective being influenced by its assigned weight. 

Further, in our method, each perspective's initial state is identical to that of the original model, ensuring that initially, all perspectives behave similarly. We then introduce noise into these perspectives and fine-tune them aiming to surpass the original model's performance (see next subsection). This strategy is effective because: (a) starting from the same initial state as the original model and then introducing variability through noise encourages the perspectives to adapt and improve beyond their starting point, and (b) it starts from a pre-trained model and, thus, can skip this step. Our experiments demonstrated that this method not only overcomes the limitations of the previous two methods but also leads to perspectives that are better than their original state, enhancing the overall model's performance. Equation \ref{eq:temp_eq} describes the mechanism in detail and equation \ref{eq:final_eq} sums it up as the final output.

\subsection*{Training Method and Noise Addition}

\subsubsection*{Focused training on temporal components}

In our approach, we specifically targeted the training of the temporal components of the model. This decision was made to minimize the risk of overfitting the model to the data. By concentrating on the temporal aspects, we ensure that the core functionality of the model remains robust while optimizing its ability to process and interpret temporal information. To this end, we fine-tune the architecture by freezing the original parameters of the model and train the temporal components. We further detail this process in the next section.

\subsubsection*{Strategic noise addition to the perspective selector}

A notable aspect of our training process involves the addition of noise to help differentiate between the multiple perspectives.
Empirical observations from our experiments indicated that adding noise to the perspective aggregator (the ``Linear'' layer component in Figure~\ref{fig:our_combination}) results in more favorable outcomes compared to introducing noise directly to the temporal values of the perspectives. This observation led us to hypothesize that adding noise to the temporal values might lead to the loss of valuable information. Therefore, by introducing noise to the linear layer responsible for selecting perspectives, we can enhance the model's performance without compromising the integrity of the temporal data. The initial noise applied to this linear layer is  calibrated with a standard deviation of 0.01 and a mean of 0, ensuring a subtle yet effective alteration in the model's learning process.

\begin{table}[h]
\caption{\footnotesize Number of parameters compared with the original RWKV-v4 model.}
\vskip 0.15in
\centering
\begin{tabular}{lll}
\hline
\Tstrut
RWKV-v4 Size & Our Approach & Increase \% \Bstrut\\
\hline
\Tstrut
$1.6934 \times 10^8$ & $1.6948 \times 10^8$  & $0.08\%$      \\
$4.3039 \times 10^8$   & $4.3077 \times 10^8$  & $0.09\%$         \\
$1.5151 \times 10^9$ & $1.5158 \times 10^9$  & $0.04\%$     \Bstrut\\
\hline
\end{tabular}
\label{table:number_parameters}
\vskip -0.1in
\end{table}

\subsection*{Experimental Setup}

As mentioned in the previous section, the temporal perspectives were initialized with the weights of the original RWKV-v4 model.
We employed a random search strategy to select hyperparameters over several training iterations, optimizing them within the confines of our available hardware resources. 
Due to hardware limitations, we constrained our batch size to 2 and concentrated our training on the novel aspect of our architecture --the temporal perspectives-- without retraining the entire model.

We conducted our experiments across three RWKV-v4 model sizes: 169M, 430M, and 1.5B parameters. The smaller models underwent training for 8 mini-epochs, with each comprising 16,000 contexts, amounting to roughly 131 million tokens. The largest model, with 1.5 billion parameters, was trained for 8 mini-epochs, each with 48,000 contexts chosen at random, totaling approximately 394 million tokens. Note that in the context of leading models trained on tens of trillions of tokens, the number of tokens we use for training is low. Initially, our architecture's development and testing phases were conducted on a server equipped with a single RTX 4090 GPU. Then, we transitioned to using a single NVIDIA A100 GPU for our final training rounds. Training one mini-epoch on the smallest model took approximately 30 minutes, while the medium-sized model required around 1 hour per mini-epoch. Given the larger model's increased complexity and the three times larger dataset, completing one mini-epoch required close to 6 hours of training time. This leads to a total of 2 days for a full training of the 1.5B model, which we ran 3 time for each ablation study.

For model optimization, we adopted an exponential learning rate decay strategy, similar to the one in the original RWKV-v4 model. Our learning rate started at a peak of 3e-5 and decreased towards a minimum value of 1e-5. 

We evaluated our model's performance using the EleutherAI Evaluation Harness \cite{eval-harness}. 

Each experiment was replicated three times using unique random seeds to assess the stability of the results.

\subsection*{Ablation Studies}
\label{sec:ablation_studies}

In this section, we examine the effects of various components and design choices within our approach. 
All ablation studies were performed on the 169M-parameter model and included four temporal perspectives (unless specified otherwise). 

\subsubsection*{Number of Perspectives}

\begin{figure*}[ht]
\vskip 0.2in
\begin{center}
\centerline{\includegraphics[width=0.8\textwidth]{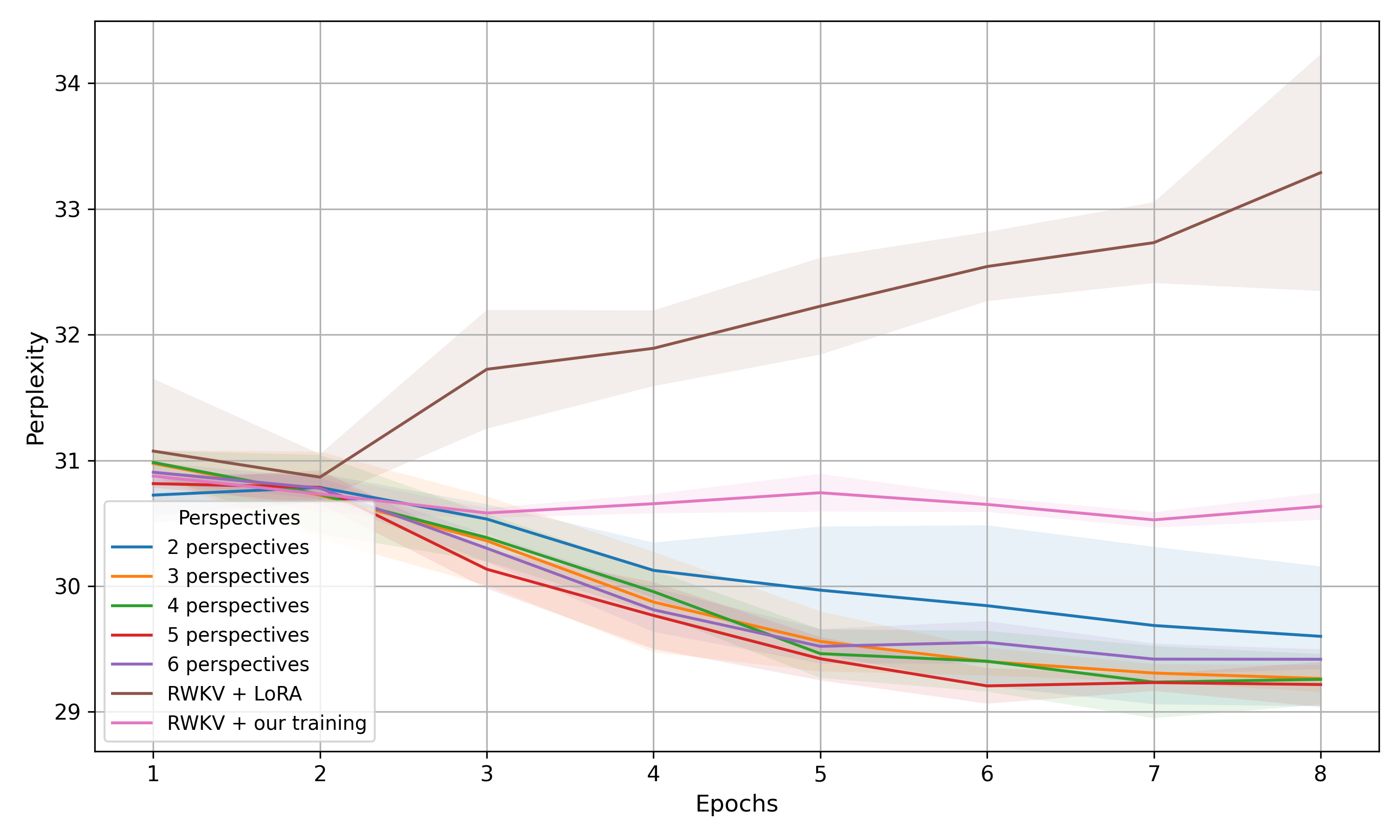}}
\caption{\footnotesize Number of perspectives impact on performance. Each line represents the mean perplexity (lower is better) across three runs, while the shaded regions indicate the standard deviation.}
\label{figure:number_perspectives}
\end{center}
\vskip -0.2in
\end{figure*}

Our first ablation study focused on the number of temporal perspectives and their influence on model performance, using the LAMBADA benchmark as a testing ground. 
The overall results are shown in Figure~\ref{figure:number_perspectives}.
This experiment revealed a positive correlation between the number of perspectives and the model's predictive capabilities, with models that featured an increased number of perspectives demonstrating superior performance. As Figure~\ref{figure:number_perspectives} shows, two to three perspectives offer the optimal balance between complexity and effectiveness, maximizing the model's understanding of the input sequence. 

Furthermore, our analysis of RWKV combined with LoRA, as well as RWKV with our training approach, indicates that the enhanced performance is attributed to the novel perspectives, rather than the additional fine-tuning on Wikipedia (already part of the pre-training set), or the additional parameters added to the model. We hypothesize that LoRA's comparatively lower performance could be due to the model being overly specialized towards Wikipedia content.

\subsubsection*{Perspective Aggregation Strategies}

\begin{table*}[h]
\caption{\footnotesize Performance of the various perspective aggregation strategies on two datasets.}
\vskip 0.15in
\centering
\begin{tabular}{llll}
\hline
\Tstrut
Data Set & Simple Average & Linear & Our Approach \Bstrut\\
\hline
\Tstrut
ARC-Easy acc & $47.32\pm{0.05}\%$  &  $45.20\pm{0.38}\%$   &  \textbf{47.42}$\pm{0.27}\%$      \\
LAMBADA ppl   & $30.44\pm{0.13}$    &  $37.39\pm{1.12}$    &  \textbf{29.25}$\pm{0.24}$        \\
LAMBADA acc   & $33.17\pm{0.17}\%$  &  $32.72\pm{1.05}\%$   &  \textbf{34.11}$\pm{0.15}\%$      \Bstrut\\
\hline
\end{tabular}
\label{table:aggregation_ablation}
\vskip -0.1in
\end{table*}

\begin{table*}[h!]
\caption{\footnotesize Impact of noise placement on two datasets.}
\vskip 0.15in
\centering
\begin{tabular}{lll}
\hline
\Tstrut
Data Set & Noise on Temporal Information & Noise on the Linear Layer \Bstrut\\
\hline
\Tstrut
ARC-Easy acc & $47.40\small{\pm{0.25}\%}$   &  \textbf{47.42}$\small{\pm{0.27}\%}$      \\
LAMBADA ppl   & $29.34\small{\pm{0.09}}$     &  \textbf{29.25}$\small{\pm{0.24}}$        \\
LAMBADA acc   & $33.88\small{\pm{0.05}\%}$   &  \textbf{34.11}$\small{\pm{0.15}\%}$      \Bstrut\\
\hline
\end{tabular}
\vspace{-1mm}
\label{table:noise_placement}
\vskip -0.1in
\end{table*}

The second study compared the three perspective aggregation techniques introduced in Section~\ref{sec:aggregation}.
The experimental results summarized in Table~\ref{table:aggregation_ablation} favored our original aggregation method, which consistently outperformed others in terms of contributing to the overall performance gains.

\subsubsection*{Noise Placement}

We explored the addition of noise both to the temporal information and to the linear layer in our perspective aggregation component. Our experiments (Table~\ref{table:noise_placement}) consistently showed that adding noise into the linear layer led to more substantial improvements in performance, suggesting that this strategy enables the model to escape local minima more effectively and generalize better.

\subsubsection*{Perspectives Weights Applied on Tokens}

In a qualitative study on the weight distribution across temporal perspectives, our findings  revealed a distinctive pattern of model reliance on two primary perspectives (see Figure~\ref{fig:example_weights}). These perspectives demonstrated an adaptive behavior, selectively prioritizing parts of the input sentence based on contextual relevance.

\section*{Limitations}

In identifying the limitations of our study, several key aspects emerged. 
The integration of multiple temporal perspectives in the RWKV architecture increases inference time complexity proportionally with the number of perspectives while maintaining the same memory utilization. However, since the perspectives are independent of each other, an efficient implementation could parallelize them retaining the original RWKV inference efficiency. Additionally, the back-propagation process requires more memory and time proportional to the added perspectives.

Our methodology, constrained by computational resources, utilized a smaller batch size of 2, possibly limiting the full potential of our approach. We anticipate that a larger batch size or training a model from scratch with our method could lead to improved outcomes, but these hypotheses remain untested due to our resource limitations.

\section*{Accessibility}

Our approach to integrating multiple temporal perspectives within RNN architectures has the potential to democratize access to recent advancements in large language models (LLMs). By enhancing the efficiency of these models, we may enable populations with limited computational resources to harness the power of advanced NLP tools.

Furthermore, we have selected a color scheme that prioritizes accessibility, ensuring the visuals are clear and discernible to individuals with color vision deficiencies. This inclusive approach reflects our commitment to making our research accessible to a wider audience, including those with varying visual abilities.

\section*{Impact Statement}

In our paper on enhancing Transformer RNNs with multiple temporal perspectives, we recognize the potential broader impacts of our work in the field of Machine Learning. While our primary objective is technical advancement, we are mindful of the ethical aspects and societal consequences of our research. Improved machine learning models, like the one we propose, could significantly impact areas such as data privacy, algorithmic bias, and automation in various industries. We believe it's important to consider these impacts as part of responsible AI development. Although our work does not directly address these ethical issues, we encourage ongoing dialogue and research in these areas to ensure the benefits of AI advancements are balanced with societal well-being and ethical considerations.




\bibliography{paper}

\end{document}